# CANDID: Robust Change Dynamics and Deterministic Update Policy for Dynamic Background Subtraction

Murari Mandal, Prafulla Saxena, Santosh Kumar Vipparthi

Department of Computer Science & Engineering
MNIT Jaipur, India
murari023@ieee.org, anshusaxena1991@gmail.com,
skvipparthi@mnit.ac.in

Subrahmanyam Murala

Computer Vision and Pattern Recognition Lab.
Department of Electrical Engineering
IIT Ropar, India
subbumurala@iitrpr.ac.in

*Abstract*—Background subtraction in video provides the preliminary information which is essential for many computer vision applications. In this paper, we propose a sequence of approaches named CANDID to handle the change detection problem in challenging video scenarios. The CANDID adaptively initializes the pixel-level distance threshold and update rate. These parameters are updated by computing the change dynamics at a location. Further, the background model is maintained by formulating a deterministic update policy. The performance of the proposed method is evaluated over various challenging scenarios such as dynamic background and extreme weather conditions. The qualitative and quantitative measures of the proposed method outperform the existing state-of-the-art approaches.

## I. Introduction

Change detection is one of the preliminary task in numerous computer vision applications such as behavior analysis, traffic monitoring, video synopsis, action recognition, visual surveillance, anomaly detection and object tracking. Background subtraction is an effective approach to detect the relevant changes by segmenting the video frames into foreground and background regions. Moving object detection in videos with dynamic background changes, illumination variations and challenging environmental conditions is a challenging task due to the fluctuation and noise in the background appearance. These dynamic changes affect the accuracy of a background subtraction technique.

One of the seminal work in background subtraction named ViBe was proposed by [1], where the authors proposed three important background model update strategies: random sample replacement, memoryless update policy, spatial diffusion via background sample propagation. They further used a constant threshold and static update rate for foreground detection and background model maintenance. Hofmann et al. [2] introduced dynamic controllers to update the per-pixel decision thresholds and learning rates which was further improved by St-Charles et al. [3] to propose a more pervasive change detection technique. The SuBSENSE [3] computes the pixel-level spatiotemporal feature descriptor LBSP [4], color channel intensity and incorporates the adaptive feedback information to perform background subtraction. The adaptive feedback mechanism continuously monitors the model fidelity and segmentation entropy to update the decision thresholds, learning rates and background samples. All these methods use the random sample update policy for background model maintenance. According to Charles et al. [3], updating the samples randomly ensures the presence of long-term and short-term history of background representation in the background model. However, this approach gives equal importance to all the background samples and thereby, both relevant and irrelevant samples have equal probability to be updated. This leads to insufficient or improper update of the background samples which is a common reason for unsatisfactory results in sample-based approaches.

Motivated by the preceding considerations, in this paper, we propose a new background subtraction technique which employs a deterministic model update policy based on the observation of recent pixel history behavior. Moreover, in order to minimize dependence on manual parameter tuning for different visual scenarios, we designed an adaptive parameter initialization and maintenance scheme.

The pixel-wise adaptive decision threshold and update rate are decided using information from multiple sources: first, the mean of temporal gradients (mTG) is computed in the parameter initialization phase and second, the pixel-level change dynamics (CD) is calculated at each newly observed frame. We maintain a recent pixel history model and update the background samples based on the current pixel intensity deviation from the mean of recent pixel history model. This approach helps in avoiding erroneous updating of background samples by stochastic selection. The proposed method works very well in dynamic background and bad weather conditions due to the computation of pixel-level change dynamics. A block diagram representation of the proposed method is shown in Fig. 1. The evaluation of the proposed method is done by computing the performance measures in various challenging videos selected from CDNet 2014 dataset [6].

The rest of the paper is organized as follows: the related background subtraction techniques are discussed in Section II.

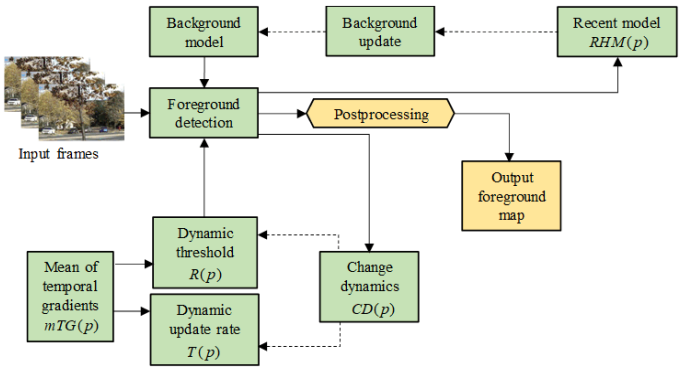

Fig. 1. Block diagram of the proposed method. In this context, $mTG(p)$ represents the pixel-level average subtracted values, $CD(p)$ is the change dynamics computed at every incoming frame, $RHM(p)$ is the recent history model, $R(p)$ contains the dynamic threshold for foreground decision making and $T(p)$ is the update rate for the background pixels. Both $R(p)$ and $T(p)$ are controlled by $mTG(p)$ and $CD(p)$ respectively.

We then describe our proposed method in Section III. Section IV discusses the experimental results and comparison of the proposed method with other state-of-the-art techniques. We conclude our work in Section V.

## II. RELATED WORK

In background subtraction, two primary tasks are to extract relevant features from the image sequences and design a robust background modelling technique. In addition to the low-level image features, i.e., grayscale, color intensity and edge magnitudes [1, 2, 5-8], specific feature descriptors can be designed for enhanced performance [3-4]. The background modelling techniques in the literature are loosely categorized into parametric [10-13] and non-parametric [14-20] techniques. A detailed classification of background modelling techniques can be found in [9]. Stauffer and Grimson [10] developed one of the most popular pixel-level parametric method called Gaussian Mixture Models (GMM), to model the statistical distribution of intensities at each pixel location. The background models are updated using the Expectation Maximization (EM) algorithm under Gaussian distribution and pixel classification is done based on matching the related Gaussian distribution with the background model. KaewTraKulPong and Bowden [11] proposed to use adaptive mixture models for updating the background model alongside fast initialization. Zivkovic [12] and Varadarajan et al. [13] further improved upon the adaptive GMM with variable parameter selection and spatial mixture of Gaussians.

The parametric methods may be susceptible to volatile/high-frequency changes in the video. In this regard, a non-parametric approach using kernel density estimation (KDE) was presented by [14] to approximate multimodal distribution. These KDE methods may become computationally expensive and require more memory. Some of these shortcomings were addressed in probability distribution function [15] based methods. In [16], the authors proposed a consensus-based method which stores a collectionof background samples at each pixel and updates the samples through first-in-first-out policy. However, suchupdate policy doesn't reflect the background behavior in real life video sequences. Barnich and Droogenbroeck [1] introduced ViBe to address some of the above-mentioned problems. They initialized the background model using only one frame and designed three very effective model update strategies. ViBe outperforms the aforementioned techniques due to its ability to capture the history of short and long-term background representation at pixel-level via a random sample update policy. Hofmann et al. [2] proposed a Pixel-Based Adaptive Segmenter (PBAS) to segment foreground regions by introducing dynamic controllers for the decision thresholds and learning rates. In [3], the authors designed a more sophisticated algorithm SuBSENSE for background subtraction based on adaptive feedback mechanism and spatiotemporal feature descriptors. Jiang and Lu [5] used a weight-sample-based method for foreground detection which utilizes few samples with variable weights to achieve effective change detection. Sajid et al. [19-20] created multiple background models based the fusion of RGB and YCbCr color models to estimate the foreground/ background probability. Neural networks based self-organizing background subtraction [21-22], motion modelling using graph cut and optical flow [23] and physics-based change detection [24] are some other interesting methods proposed by the researchers to solve the problems in motion detection. Further, Bianco et al. [25] conducted experiments to combine various change detection algorithms to improve the accuracy of the background subtractor.

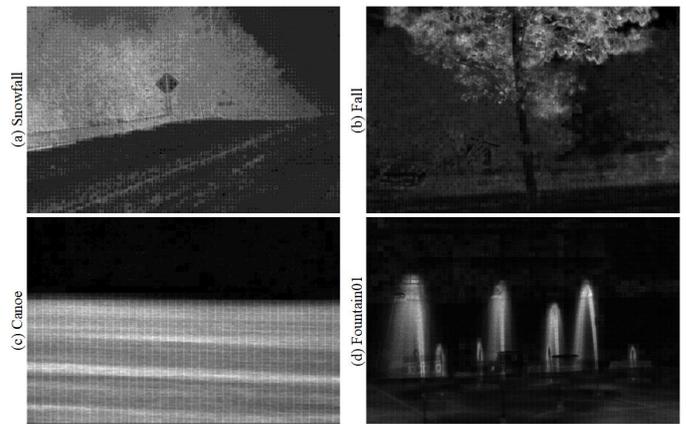

Fig. 2. The mean of temporal gradients (mTG) after initialization using initial $F_n$ frames for (a) Snowfall, (b) Fall, (c) Canoe and (d) Fountain01 videos.

## III. PROPOSED METHOD

The detailed description of the proposed method is given in the following four steps: adaptive parameter initialization, background model, change dynamics for foreground detection and deterministic update policy. For simplicity, we refer the proposed method as the robust ChANge DynamIcs and Deterministic Update Policy (CANDID).

### A. Adaptive parameter initialization

The parameter initialization is one of the distinguishable aspect of the proposed method compared to state-of-the-art techniques. In the literature, the distance threshold and learning rate are selectively initialize by the user. Whereas, in

CANDID, these parameters are initialized adaptively by computing the mean of temporal gradients (mTG). Let $I_k$ be a frame of size $P \times Q$ in a video stream $\{I_k\}_{k=1}^{V}$. The pixel coordinates of image $I_k$ is represented as $I_k(a,b) \forall (a \in [1,P], b \in [1,Q])$ and $V$ is the length of the video. Then the AD can be computed using Eq. (1) as below,

$$mTG(a,b) = \frac{1}{(F_n - 1)} \sum_{k=1}^{F_n - 1} |I_k(a,b) - I_{k+1}(a,b)| \quad (1)$$

where $F_n$ is the total number initial frames selected for parameter initialization.

The main aim of the mTG is to locate the stable and unstable regions. Thus, the initial parameters are adaptively initialized using Eq. (2) and Eq. (3). This adaptive initialization is not present in the state-of-the-art methods [1-4].

$$R_0(a,b) = mTG(a,b) + \alpha \quad (2)$$

$$T_0(a,b) = \beta \times \frac{1}{(1 + mTG(a,b))^2} \quad (3)$$

where $R_0(a,b)$ is the adaptive distance threshold, $T_0(a,b)$ is the adaptive update rate, $\alpha$ and $\beta$ are the offset parameters. The linear relation between $R_0(a,b)$ and $mTG(a,b)$ states that, the probability of a pixel being classified as foreground is higher when there is substantial deviation from the $mTG(a,b)$. To incorporate the motion entropy into the background model, the update rate is lower for the pixel where the intensity variation is higher as compared to the pixel with minimal variation. This shows an exponential relation between $T_0(a,b)$ and $mTG(a,b)$.

### B. Background model

After parameter initialization, the next task is to initialize the background model ($BM$) as defined in Eq. (4).

$$BM(a,b) = \{s_i(a,b)\}_{i=F_n+1}^{j=F_n+N} \quad (4)$$

where $s_i$ is the background sample at index position $i$ and $N$ is the total no. of samples. In addition, a recent history model (RHM) is generated by keeping the last five-pixel intensities as samples. The RHM is computed in Eq. (5).

$$RHM(a,b) = \{I_i(a,b)\}_{i=j-5}^{j} \quad (5)$$

### C. Change dynamics for foreground detection

The foreground segmentation rules are defined based on the pixel-level change dynamics (CD). The aim of the CD is to estimate the motion pattern at a location based on the background model proximity to the current intensity value. The authors of SuBSENSE identified the background dynamics at each location by computing recursive moving average in continuous manner whereas, in our work, the CD is computed based on the current background sample distances. A higher value of CD represents frequent fluctuation of intensity values at that position. Hence, the dynamic background pixels in the video can be identified. The effect of CD over the dynamic background videos *fountain01* and *canoe* [6] is shown in Fig. 3. From Fig. 3, it is clear that, the CD is able to distinguish between the temporal variations of the background and foreground pixels. The CD at frame $k$ can be computed as follows,

$$CD_k(a,b) = \frac{1}{2 \times L^2} MP_k(a,b) \times [med(\{FS_{k,i}(a,b)\}_{i=1}^{N/2}) + med(\{FS_{k,i}(a,b)\}_{i=N/2+1}^{N})] \quad (6)$$

where $L = 255$ and $MP_k(a,b) = \frac{1}{N} \sum_{i=1}^{N} \{DB_k(a,b)\}_i$. The $DB_k(a,b)$ and $FS_k(a,b)$ can be computed using Eq. (7) and Eq. (8) respectively.

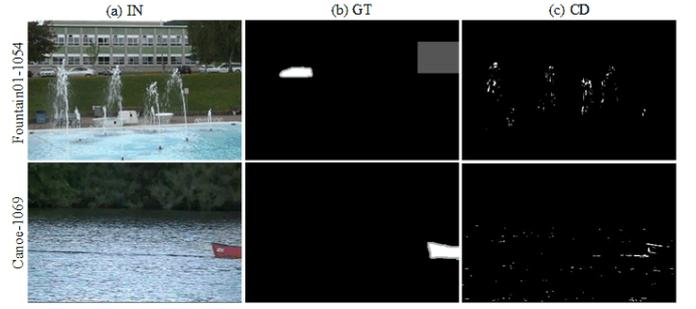

Fig. 3. Change dynamics in dynamic background videos to identify the unstable pixel locations

$$DB_k(a,b) = \{|I_k(a,b) - BM_i(a,b)|\}_{i=1}^{N} \quad (7)$$

$$FS_k(a,b) = sort(DB_k(a,b)) \quad (8)$$

After the CD is calculated, the decision threshold $R$ is adaptively updated to detect the foreground with minimal false detections. A pixel with higher CD value has greater probability to be part of the dynamic region and therefore the distance threshold for this pixel is increased. On the other hand, if the CD value is negligible, then the initialized distance threshold is sufficient for accurate detection. The distance threshold $R_k(a,b)$ at frame $k$ is computed using Eq. (9).

$$R_k(a,b) = \begin{cases} R_0(a,b) + \gamma, & \text{if } CD_k(a,b) > \xi \\ R_0(a,b), & \text{otherwise} \end{cases} \quad (9)$$

where the $\gamma$ is the offset parameter to adjust the $R_k(a,b)$ and $\xi$ is the degree of change in the CD. The foreground detection is performed as presented in Eq. (10).

$$F_k(a,b) = \begin{cases} 0, & \text{if } X_k(a,b) < \#_{min} \\ 1, & \text{otherwise} \end{cases} \quad (10)$$

where $\#_{min}$ is the minimum no. of matches required to label a pixel as background. In our experiments, we set the parameter $\#_{min} = 2$. The $X_k(a,b)$ can be computed through Eq. (11).

$$X_k(a,b) = \#\{DB_{k,i}(a,b) < R_k(a,b), \forall N\}_i \quad (11)$$

The update rate $T$ is also updated based on the CD value. Since, the background model is updated with $1/T$ probability, it is quite clear that when the CD value is higher, the update rate is decremented to ensure higher probability of updating background samples and vice-versa. The update rate $T_k(a,b)$ is computed using Eq. (12).

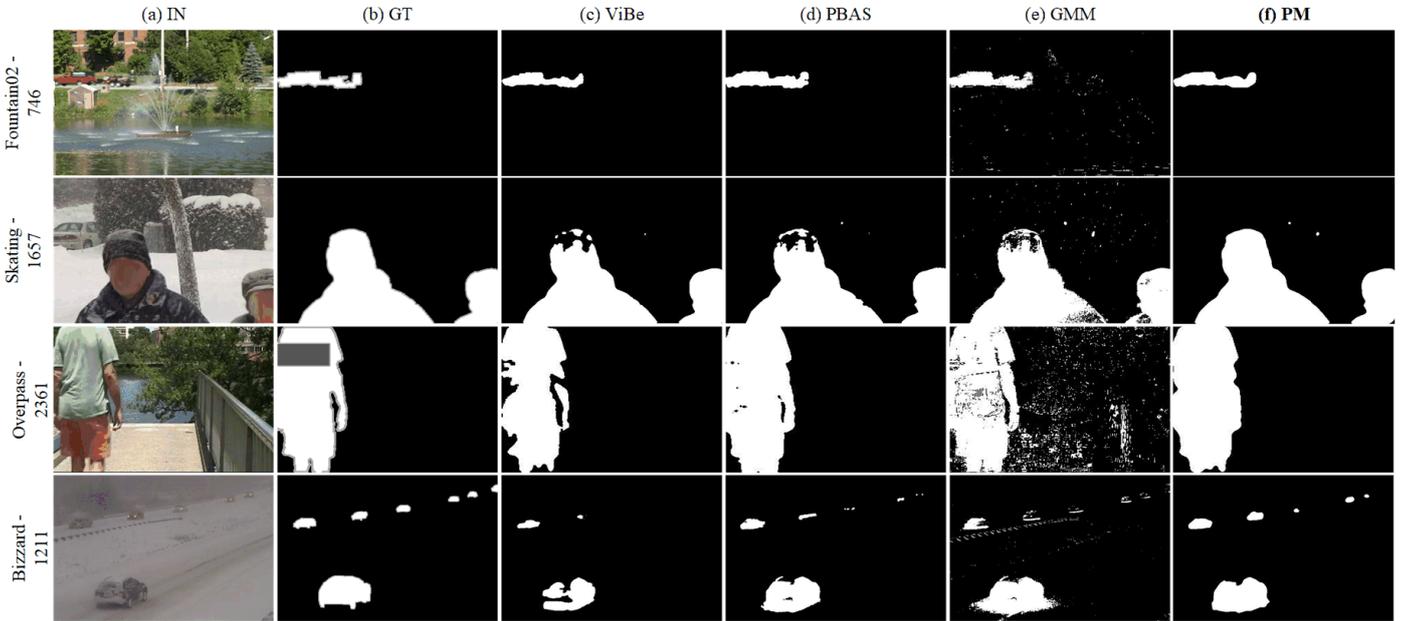

Fig. 4. Foreground detection results of the proposed method and other state-of-the-art methods

$$T_k(a,b) = \begin{cases} 2, & \text{if } CD_k(a,b) > \xi \\ K, & \text{otherwise} \end{cases} \quad (12)$$

where $K = 1/CD_k(a,b)$, and it is bound within the interval $[2,300]$.

### D. Deterministic update policy

The background model updating procedure is very crucial for developing a robust background subtraction technique. Without a proper update mechanism in place, the performance of the method will deteriorate in challenging video scenarios. In our work, we update the background model based on a deterministic policy. The $RHM_k(a,b)$ is utilized to determine the update procedure in the $BM_k(a,b)$. For this purpose, we first compute the $RDist_k(a,b)$ as follows,

$$RDist_k(a,b) = mean(RHM_{k-1}(a,b)) - I_k(a,b) \quad (13)$$

If $RDist_k(a,b) > 0 \,\&\&\, F_k(a,b) == 0$, then the background sample having minimum distance is replaced by the current pixel value. If $RDist_k(a,b) < 0 \,\&\&\, F_k(a,b) == 0$, then the background sample with maximum distance is replaced by the current pixel value. This mechanism identifies the samples which may no longer be relevant in the background model and thus deterministically replaces the irrelevant background sample with the current pixel. Further, the neighborhood pixel models of all the background pixels are also updated in similar manner to ensure spatial diffusion [1]. We plan to make the source code of CANDID available after the acceptance of the paper.

## IV. EXPERIMENTAL RESULTS

The effectiveness of the proposed method is validated in various challenging visual scenarios including dynamic and noisy background changes. In this paper, ten videos are selected from the dynamic background and bad weather

TABLE I

EVALUATION RESULTS OF THE PROPOSED METHOD ON THE DYNAMIC BACKGROUND AND BAD WEATHER VIDEOS FROM CDNET 2014 DATASET

| Video | Pr | Re | FM | Sp | PWC |
|---|---|---|---|---|---|
| **Blizzard** | 0.93 | 0.80 | 0.87 | 1.00 | 0.31 |
| **Skating** | 0.96 | 0.88 | 0.92 | 1.00 | 0.80 |
| **Snowfall** | 0.75 | 0.77 | 0.78 | 1.00 | 0.39 |
| **wetSnow** | 0.83 | 0.82 | 0.83 | 1.00 | 0.45 |
| **Boats** | 0.93 | 0.50 | 0.66 | 1.00 | 0.34 |
| **Canoe** | 0.99 | 0.81 | 0.91 | 1.00 | 0.70 |
| **Fall** | 0.67 | 0.97 | 0.81 | 0.99 | 0.91 |
| **fountain01** | 0.38 | 0.87 | 0.55 | 1.00 | 0.13 |
| **fountain02** | 0.96 | 0.86 | 0.92 | 1.00 | 0.04 |
| **Overpass** | 0.97 | 0.84 | 0.92 | 1.00 | 0.25 |
| **Avg.** | **0.84** | **0.81** | **0.82** | **1.00** | **0.43** |

categories of CDNet 2014 dataset [6]. The proposed method adaptively computes the initial parameters using the first $B$ frames. These parameters are learned with time based on the change dynamics (CD) at each pixel location. The performance of the proposed method is measured in terms of *Precision (Pr), Recall (Re), F-Measure (FM), Specificity (Sp)* and *Percentage of Wrong Classification (PWC)*. These metrics are calculated based on the true positives, true negatives, false positives and false negatives. All the performance measures were computed as defined in [6] to ensure consistency with the comparative results of the existing approaches.

### A. Model parameters

Since, our method is designed to address the dynamic background behavior and noise in the video sequences, we have conducted numerous experiments with different parameters and selected the following optimal values: $F_n = 300$, $\alpha = 10$, $\beta = 50$, $N = 30$, $\gamma = 10$, $\xi = 0.1$. In this

TABLE II

COMPARATIVE CHANGE DETECTION PERFORMANCE OF THE PROPOSED METHOD AND EXISTING STATE-OF-THE-ART METHODS BASED ON FM AND PWC OVER THE DYNAMIC BACKGROUND AND BAD WEATHER VIDEOS FROM CDNET 2014 DATABASE

| Methods | Metric | blizzard | skating | snowFall | wetSnow | boats | canoe | fall | fount01 | fount02 | overpass | Avg. |
|---|---|---|---|---|---|---|---|---|---|---|---|---|
| GMM_Grim [10] | FM | 0.83 | 0.86 | 0.76 | 0.61 | 0.73 | 0.88 | 0.44 | 0.08 | 0.80 | 0.87 | 0.69 |
|  | PWC | 0.36 | 1.22 | 0.37 | 0.98 | 0.35 | 0.82 | 4.05 | 1.60 | 0.09 | 0.33 | 1.02 |
| GMM_Zivk [12] | FM | 0.80 | 0.84 | 0.74 | 0.56 | 0.26 | 0.64 | 0.32 | 0.05 | 0.58 | 0.67 | 0.55 |
|  | PWC | 0.40 | 1.34 | 0.36 | 1.02 | 1.91 | 3.02 | 5.61 | 1.84 | 0.23 | 0.98 | 1.67 |
| KDE [14] | FM | 0.54 | 0.80 | 0.41 | 0.12 | 0.03 | 0.18 | 0.08 | 0.01 | 0.19 | 0.24 | 0.26 |
|  | PWC | 0.81 | 2.01 | 1.24 | 12.32 | 35.14 | 33.19 | 35.45 | 13.88 | 1.72 | 8.33 | 14.41 |
| VIBE [1] | FM | 0.53 | 0.71 | 0.66 | 0.55 | 0.22 | 0.75 | 0.42 | 0.09 | 0.65 | 0.68 | 0.53 |
|  | PWC | 0.75 | 2.95 | 0.42 | 0.91 | 1.61 | 1.80 | 3.30 | 0.76 | 0.14 | 0.82 | 1.35 |
| PBAS [2] | FM | 0.82 | 0.89 | 0.73 | 0.72 | 0.21 | 0.40 | 0.89 | 0.59 | 0.90 | 0.66 | 0.68 |
|  | PWC | 0.38 | 1.03 | 0.37 | 0.61 | 0.56 | 2.67 | 0.40 | 0.10 | 0.04 | 0.70 | 0.69 |
| LOBSTER [4] | FM | 0.47 | 0.78 | 0.65 | 0.53 | 0.58 | 0.93 | 0.25 | 0.16 | 0.83 | 0.70 | 0.59 |
|  | PWC | 0.81 | 2.08 | 0.42 | 0.89 | 0.37 | 0.49 | 8.90 | 0.67 | 0.07 | 0.99 | 1.57 |
| SuBSENSE [3] | FM | 0.85 | 0.89 | 0.88 | 0.80 | 0.69 | 0.79 | 0.87 | 0.75 | 0.94 | 0.86 | 0.83 |
|  | PWC | 0.32 | 0.95 | 0.19 | 0.46 | 0.31 | 1.22 | 0.47 | 0.05 | 0.02 | 0.35 | 0.43 |
| UBSS1 [21] | FM | 0.87 | 0.92 | 0.85 | 0.56 | 0.90 | 0.93 | 0.57 | 0.52 | 0.92 | 0.90 | 0.80 |
|  | PWC | 0.29 | 0.72 | 0.25 | 1.53 | 0.11 | 0.44 | 2.01 | 0.06 | 0.03 | 0.25 | 0.57 |
| UBSS2 [20] | FM | 0.87 | 0.92 | 0.85 | 0.48 | 0.90 | 0.93 | 0.57 | 0.52 | 0.92 | 0.90 | 0.79 |
|  | PWC | 0.29 | 0.72 | 0.25 | 2.05 | 0.11 | 0.44 | 2.01 | 0.06 | 0.03 | 0.25 | 0.62 |
| Spectral-360 [24] | FM | 0.78 | 0.92 | 0.76 | 0.65 | 0.69 | 0.88 | 0.90 | 0.47 | 0.92 | 0.81 | 0.78 |
|  | PWC | 0.43 | 0.75 | 0.34 | 0.94 | 0.30 | 0.78 | 0.37 | 0.17 | 0.03 | 0.46 | 0.46 |
| IUTIS-1 [25] | FM | 0.67 | 0.71 | 0.76 | 0.55 | 0.32 | 0.41 | 0.18 | 0.04 | 0.74 | 0.83 | 0.52 |
|  | PWC | 0.59 | 3.44 | 0.33 | 1.32 | 2.02 | 9.90 | 14.83 | 3.39 | 0.14 | 0.48 | 3.64 |
| IUTIS-2 [25] | FM | 0.63 | 0.89 | 0.76 | 0.73 | 0.59 | 0.71 | 0.30 | 0.07 | 0.89 | 0.88 | 0.65 |
|  | PWC | 0.64 | 0.95 | 0.33 | 0.59 | 0.48 | 1.96 | 7.26 | 1.98 | 0.05 | 0.30 | 1.45 |
| RMoG [13] | FM | 0.61 | 0.79 | 0.77 | 0.64 | 0.83 | 0.94 | 0.67 | 0.20 | 0.87 | 0.90 | 0.72 |
|  | PWC | 0.66 | 1.73 | 0.33 | 0.77 | 0.21 | 0.44 | 1.23 | 0.36 | 0.06 | 0.25 | 0.60 |
| SC-SOBS [21] | FM | 0.59 | 0.89 | 0.65 | 0.51 | 0.90 | 0.95 | 0.28 | 0.12 | 0.89 | 0.88 | 0.67 |
|  | PWC | 0.68 | 1.04 | 0.43 | 1.23 | 0.13 | 0.34 | 8.35 | 0.93 | 0.05 | 0.34 | 1.35 |
| BingWang [17] | FM | 0.73 | 0.89 | 0.78 | 0.67 | 0.85 | 0.93 | 0.63 | 0.77 | 0.93 | 0.95 | 0.81 |
|  | PWC | 0.50 | 0.97 | 0.30 | 0.78 | 0.19 | 0.53 | 1.97 | 0.04 | 0.03 | 0.14 | 0.55 |
| CP3 [15] | FM | 0.68 | 0.90 | 0.74 | 0.80 | 0.54 | 0.91 | 0.63 | 0.17 | 0.64 | 0.77 | 0.68 |
|  | PWC | 1.00 | 1.00 | 0.43 | 0.53 | 0.87 | 0.61 | 1.18 | 0.61 | 0.16 | 0.52 | 0.69 |
| AAPSA [22] | FM | 0.84 | 0.85 | 0.78 | 0.69 | 0.76 | 0.89 | 0.75 | 0.44 | 0.36 | 0.82 | 0.72 |
|  | PWC | 0.33 | 1.31 | 0.34 | 0.64 | 0.24 | 0.72 | 0.79 | 0.11 | 0.71 | 0.41 | 0.56 |
| EFiC [7] | FM | 0.73 | 0.92 | 0.86 | 0.57 | 0.36 | 0.36 | 0.72 | 0.23 | 0.91 | 0.88 | 0.66 |
|  | PWC | 0.50 | 0.74 | 0.21 | 1.58 | 0.53 | 2.88 | 1.26 | 0.47 | 0.04 | 0.32 | 0.85 |
| C-EFiC [8] | FM | 0.76 | 0.90 | 0.87 | 0.62 | 0.37 | 0.34 | 0.56 | 0.27 | 0.93 | 0.90 | 0.65 |
|  | PWC | 0.45 | 0.94 | 0.20 | 1.20 | 0.50 | 2.91 | 2.43 | 0.37 | 0.03 | 0.26 | 0.93 |
| Graphcut [23] | FM | 0.90 | 0.92 | 0.88 | 0.87 | 0.57 | 0.12 | 0.72 | 0.08 | 0.91 | 0.84 | 0.68 |
|  | PWC | 0.24 | 0.76 | 0.20 | 0.31 | 0.52 | 52.01 | 1.20 | 1.10 | 0.04 | 0.40 | 5.68 |
| MultSpat [18] | FM | 0.71 | 0.62 | 0.71 | 0.57 | 0.48 | 0.89 | 0.41 | 0.14 | 0.82 | 0.84 | 0.62 |
|  | PWC | 0.52 | 4.78 | 0.37 | 0.97 | 0.70 | 0.83 | 4.27 | 0.51 | 0.08 | 0.43 | 1.35 |
| **PM** | **FM** | **0.87** | **0.92** | **0.78** | **0.83** | **0.67** | **0.91** | **0.81** | **0.55** | **0.92** | **0.92** | **0.82** |
|  | **PWC** | **0.31** | **0.80** | **0.39** | **0.45** | **0.34** | **0.70** | **0.91** | **0.13** | **0.04** | **0.25** | **0.43** |

experiment, we have applied a 7 by 7 median filter as a postprocessing technique.

*B. Results and discussions*

The detailed foreground detection results of the proposed method as compared to the state-of-the-art methods ViBe, PBAS and GMM are shown in Fig. 4. From Fig. 4, it is clear that, the proposed method is able to capture the foreground regions accurately even in the dynamic background (i.e. *fountain02* and *overpass*) and challenging bad weather conditions (i.e. *skating* and *blizzard*). In comparison to ViBe and PBAS, the proposed method is more accurately detecting the foreground regions and also removing the false positive noise in the background regions.

Quantitative measures of the proposed method on various video sequences are given in Table I. The robustness of the algorithm is measured by larger values of *Precision, Recall, Sp and FM* and smaller values of *PWC*.

We have compared the proposed method with the recent state-of-the-art background subtraction methods [6, 26] based on the FM and PWC, which are generally accepted as good indicators of overall performance. In Table 2, the comparative results of the proposed method (PM) along with twenty other approaches are given. The proposed method achieves an average of 0.82, 0.43 in terms of FM, PWC respectively. From Table II, it is clear that, the performance of the proposed method is better as compared to the recent techniques in the literature. It also achieves PWC rate equal to SuBSENSE which is based on color and spatiotemporal feature descriptors. Whereas, our algorithm only uses the grayscale intensity of the image for all the computations. In the overpass and skating videos, the PM obtains the highest and joint-highest FM out of all the comparative methods.

*C. Procesesing Speed*

The experiments were carried out on a computer with Intel Core i7-6500 processor (@2.50GHz) and 8GB DDR3 memory. The program was implemented with MATLAB without any parallelization. The runtime of the PM for $240 \times 320$ image sequences is more than 4 fps which is better than SuBSENSE which has below 2 fps runtime speed. The number of samples used in the proposed method in 30 which is significantly lower compared to the 50 samples used in SuBSENSE. Furthermore, the proposed method stores only the grayscale intensities which can reduce the memory consumption. Whereas, SuBSENSE stores three color intensities and per channel LBSP descriptors. Based on the above observations, the proposed method is suitable for online moving object detection.

## V. CONCLUSION

We have presented an adaptive background subtraction technique with deterministic background model update policy. The proposed method initializes the parameters in the early stage and then adaptively learns the pixel-level distance threshold, update rate using the change dynamics and initialized parameters. Further, the background model is updated deterministically based on the recent history model. It achieves significant performance improvement in the videos of dynamic background and bad weather scenarios. The qualitative and quantitative measures of the proposed method outperform the state-of-the-art approaches. The grayscale intensity-based computation makes the proposed method more suitable for real-time applications. In future, we would like to improve the proposed method to handle even more challenging scenarios like camera jitter and PTZ video categories.